\documentclass[graybox]{svmult}

\usepackage{mathptmx}       
\usepackage{helvet}         
\usepackage{courier}        
\usepackage{type1cm}        
\usepackage{makeidx}         
\usepackage{graphicx}        
\usepackage{multicol}        

\usepackage{amsmath}
\usepackage{algorithm}
\usepackage{algorithmic}
\usepackage{multirow}
\usepackage{multicol}
\usepackage{slashbox}

\makeindex             

\begin{document}


\title*{Ant Colony based Feature Selection Heuristics for Retinal Vessel Segmentation}

\author{Ahmed.H.Asad, Ahmad Taher Azar, Nashwa El-Bendary, Aboul Ella Hassaanien}
\institute{Ahmed.H.Asad \at ISSR, Computer Sciences and Information Dept., Cairo University, Cairo - Egypt, \\
Scientific Research Group in Egypt (SRGE), http://www.egyptscience.net\\ \email{ah\_assad@hotmail.com}
\and Ahmad Taher Azar \at Faculty of computers and information, Benha University, Egypt, \\
Scientific Research Group in Egypt (SRGE), http://www.egyptscience.net\\ \email{ahmad\_T\_azar@ieee.org}
\and Nashwa El-Bendary \at Arab Academy for Science,Technology, and Maritime Transport, Cairo - Egypt, \\
Scientific Research Group in Egypt (SRGE), http://www.egyptscience.net\\ \email{nashwa.elbendary@ieee.org}
\and Aboul Ella Hassanien \at Information Technology Dept., Faculty of Computers and Information, Cairo University, \\Cairo - Egypt, \\
Scientific Research Group in Egypt (SRGE), http://www.egyptscience.net\\ \email{aboitcairo@gmail.com}}
%
%
\maketitle

\abstract*{
Features selection is an essential step for successful data classification, since it reduces the data dimensionality by removing redundant features. Consequently, that minimizes the classification complexity and time in addition to maximizing its accuracy. In this article, a comparative study considering six features selection heuristics is conducted in order to select the best relevant features subset. The tested features vector consists of fourteen features that are computed for each pixel in the field of view of retinal images in the DRIVE database. The comparison is assessed in terms of sensitivity, specificity, and accuracy measurements of the recommended features subset resulted by each heuristic when applied with the ant colony system. Experimental results indicated that the features subset recommended by the relief heuristic outperformed the subsets recommended by the other experienced heuristics.}

\abstract{
Features selection is an essential step for successful data classification, since it reduces the data dimensionality by removing redundant features. Consequently, that minimizes the classification complexity and time in addition to maximizing its accuracy. In this article, a comparative study considering six features selection heuristics is conducted in order to select the best relevant features subset. The tested features vector consists of fourteen features that are computed for each pixel in the field of view of retinal images in the DRIVE database. The comparison is assessed in terms of sensitivity, specificity, and accuracy measurements of the recommended features subset resulted by each heuristic when applied with the ant colony system. Experimental results indicated that the features subset recommended by the relief heuristic outperformed the subsets recommended by the other experienced heuristics.}

\section{Introduction}

Retinal blood vessels are important structures in many ophthalmological images. The automated extraction of blood vessels in retinal images is an important step in computer aided diagnosis (CAD) and treatment of diabetic retinopathy \cite{Assad2012, Vijayakumari2012, Serrarbassa2008, Soares2006, Mendonca2006, Staal2004}, hypertension \cite{Leung2004}, glaucoma \cite{Wang2006}, obesity \cite{Mitchell2005}, arteriosclerosis and retinal artery occlusion, etc. 
These diseases often result in changes on reflectivity, bifurcations, tortuosity as well as other patterns of blood vessels. Vessel tortuosity, characterizes hypertension retinopathy \cite{Foracchia2001} and diabetic retinopathy, which usually leads to neovascularization. The diabetic retinopathy is one of the most common causes of vision defects or even blindness worldwide \cite{Morello2007}, as it spreads diabetes on the retina vessels, thus they lose blood supply that causes blindness in short time. Hence, analyzing vessel features gives new insights to diagnose the corresponding disease early.
So, if abnormal signs of diabetic retinopathy could be detected early, effective treatment before their initial onset can be performed and blindness can be prevented in more than 50\% of cases \cite{Xu2006, Jin2005}. However, manual segmentation of retinal blood vessels is a long and tedious task, which also requires training and skill. Hence, the early detection can be achieved via automatic segmentation of retinal blood vessels, which is the critical component of circulatory blood vessel analysis systems \cite{Fraz2012}.
Also, retinal blood vessels segmentation is the core stage in automated registration of two retinal blood vessels images of a certain patient to follow and diagnose disease progress at different periods of time \cite{Khan2011}. 

In this article, an improvement of previous work \cite{Assad2012} is presented by conducting a comparative study for six feature selection heuristics; namely, correlation-based feature selection (CFS), Fisher score, Gini index, Relief, sequential forward selection (SFS), and sequential backward selection (SBS). The aim of this article is to recommend the best relevant features subset from features vector consists of fourteen features that are computed for each pixel in the field of view (FOV) of retinal image in the DRIVE database \cite{Staal2004}. The six experimented feature selection algorithms are compared in terms of sensitivity, specificity, and accuracy of their recommended features subsets when applied with ant colony system (ACS) algorithm \cite{Dorigo1997}. The rest of this article is organized as follows: Section 2 describes in more details the used database, samples selection, selected features, and the ant colony system. Section 3 presents the six features selection heuristics in more details. Section 4 reports the results and experimental evaluations of the proposed approach. Finally, in Section 5, conclusions and directions for future research are discussed.


\section{Methods}
\subsection{Retinal Image Database and Samples Selection}

The Digital Retinal Images for Vessel Extraction (DRIVE) database \cite{Staal2004}, which consists of a total of JPEG 40 color fundus images; including 7 abnormal pathology cases, is used in this article. The images were obtained from a diabetic retinopathy screening program in the Netherlands. The images were acquired using Canon CR5 non-mydriatic 3CCD camera with FOV equals to 45 degrees. Each image resolution is 584*565 pixels with eight bits per color channel (3 channels). The set of 40 images was equally divided into 20 images for the training set and 20 images for the testing set. Inside both sets, for each image, there is circular field of view (FOV) mask of diameter that is approximately 540 pixels. Inside training set, for each image, one manual segmentation by an ophthalmological expert has been applied. Inside testing set, for each image, two manual segmentations have been applied by two different observers, where the first observer segmentation is accepted as the ground-truth for performance evaluation.

Images in DRIVE training set contain 569415 vessel pixels and 3971591 non-vessel pixels, where samples are needed to be selected for features computation. Since the ratio of vessel pixels to non-vessel pixels in each image and overall images is 1:7 as an average, the samples set consists of 1000 vessel pixels and 7000 non-vessel pixels from each image. So, for 20 images, it will be 160000 total samples. In the previous work \cite{Assad2012}, the samples set consisted of 100000 vessel and 100000 non-vessels pixels selected randomly from the whole training set using correlation-based feature selection (CFS) heuristic. 

\subsection{Selected Features}
The selected features are the gray-level of green channel of RGB retinal image, group of five features based on gray-level ($f_1$, $f_2$, $f_3$, $f_4$, $f_5$) and group of eight features based on Hu moment-invariants ($H_{u1}$, $H_{u2}$, $H_{u3}$, $H_{u4}$, $H_{u5}$, $H_{u6}$, $H_{u7}$, $H_{u8}$) \cite{Assad2012}. Most of the vessels segmentation approaches extracted and used the green color image of RGB retinal image for further processing, since it has the best contrast between vessels and background. 
The five gray-level based features group is presented by Marin et al. in \cite{Marin2011} and its features describe the gray-level variation between vessel pixel and its surrounding pixels. The Hu moment-invariants \cite{Hu1962} are the best shape descriptors, as they are invariant to translation, scale, and rotation change. So, they are used by the second group of eight features to describe vessels having variant widths and angles. The selected features are simple, well-discriminating between vessel and non-vessel classes and don't need to be computed at multiple scales or orientations. The features computation is more detailed in \cite{Assad2012}.

\subsection{Ant Colony System (ACS)}
For solving the travelling sales man problem of exploring the shortest path, ACS was first proposed in \cite{Dorigo1997}. It simulates the foraging behavior of real ants in nature, where multiple ants are going out in random paths when they are searching for foods. As the ant is moving, it is depositing a chemical substance, called "pheromone" on its path to attract other subsequent ants to follow that path. As the time passes, the pheromone evaporates so that the pheromone on the shorter path only remains for a longer time causing other ants to be attracted to it. Thus, the shortest path is the only path that attracts other ants to follow it.


\section{Features Selection Heuristics}
Feature selection plays an important role in building classification systems \cite{Hua2009}, as it can not only reduce the dimension of data, but also lower the computation consumption so that good classification performance can be obtained. In general, feature selection algorithms, designed with different evaluation criteria, are divided into two categories; namely \textit{filter methods}, which rank features or select a subset of them based on evaluation criterion, and \textit{wrappers methods}, which use predetermined learning algorithms (classifiers) to select those features with high prediction performance \cite{Blum1997, Talavera2005}. 
This section briefly describes the six feature selection heuristics experienced in this article; namely, \textit{correlation-based feature selection (CFS)}, \textit{Fisher score}, \textit{Gini index}, \textit{Relief}, \textit{sequential forward selection (SFS)}, and \textit{sequential backward selection (SBS)}.

\subsection{Correlation-based feature selection (CFS)}
It is a heuristic approach for evaluating the worth or merit of a subset of features \cite{Hall2000, Hall1999}. The main idea behind this selection method is that the most effective features for classification are those that are most highly correlated with the classes (intensifiers and dissipaters). At the same time, these features are the least correlated with other features. The CFS method is therefore used to choose a subset of features that highly represents these qualities. The best individual feature based on the merit metric calculated by equation (\ref{eq:ms})

\begin{equation}
\label{eq:ms}
M_s = \frac{k\bar{r}_{cf}}{\sqrt{k+k(k+1) \bar{r}_{ff}}}
\end{equation}

where $M_s$ is the heuristic merit of a features subset $S$ containing $k$ features, $\bar{r}_{cf}$  is the average feature-class correlation, and $\bar{r}_{ff}$ is the average feature-feature inter-correlation. The numerator gives an indication of how predictive a group of features are, while the denominator represents how much redundancy existed among them”.

\subsection{Fisher Score}
Fisher Score \cite{Bishop1996} is a type of supervised feature selection methods for examining the correlation between projected data samples and their class labels on each feature axis. It looks for features on which the classes are compact and far from each others. By considering the sample coordinates on the feature $f_r$, each class $\omega$, $\omega$ = 1,. . ., $c$, populated with $n_{\omega}$ labeled samples is characterized by its mean $\mu_{\omega r}$ and its variance ${\sigma}^2_{\omega r}$. Moreover, assume that $\mu_r$ is the mean of all data samples on the feature $f_r$, the Fisher score $F_r$ used to evaluate the relevance of the feature $f_r$ is defined as shown in equation (\ref{eq:fr}):

\begin{equation}
\label{eq:fr}
F_r = \frac{\sum^c_{\omega=1} n_\omega (\mu_{\omega r} - \mu_r)^2}{\sum^c_{\omega=1} n_\omega \sigma^2_{\omega r}}
\end{equation}

In order to select the most relevant features, they are sorted according to the decreasing order of their Fisher score $F_r$.

\subsection{Gini Index}
Gini index is a non-purity split method that fits sorting, binary systems, continuous numerical values, etc. \cite{Shang2007}. It is based on iterative splitting of the data samples into subsets based on their feature values and so on until gaining coherent samples subset \cite{Gini1963}. This approach belongs to the filter class and it ranks the features in ascending order, where the one with the smallest index is the more relevant. Suppose $\textit{S}$ is the set of \textit{s} samples. These samples have $\textit{m}$ different classes ($C_\textit{i}$ , $\textit{i}$ = 1,. . ., $\textit{m}$) based on its feature values. So, $\textit{S}$ can be divided into $\textit{m}$ subsets ($S_i$ , $\textit{i}$ = 1,. . ., $\textit{m}$). Then $S_i$ is the samples subset that belongs to class $C_i$ and $s_i$ is the samples number of subset $S_i$, then the Gini index of set $\textit{S}$ is computed as shown in equation (\ref{eq:gini})

\begin{equation}
\label{eq:gini}
Gini(S) = 1- \sum^m_{i=1} P^2_i
\end{equation}

where $P_i$ is the probability that any sample belongs to $C_i$ and estimating with $s_{i} \backslash s$.

\subsection{Relief}
This heuristic estimates the features according to how well their values discriminate among instances of different classes that are near each other \cite{Kira1992}. It's one of filter class that weight features. The relief operates as follows: for a user-specified number of instances \textit{m} and for each instance \textit{X}, relief searches for its two nearest neighbours; one from the same class (called nearest hit \textit{H}) and the other from a different class (called nearest miss \textit{M}). For each feature, it calculates its relevance score according to equation (\ref{eq:WA})
  
\begin{equation}
\label{eq:WA}
W(A) = W(A) - diff(A, X, H) / m +  diff(A, X, M) /m
\end{equation}

where $W(\textit{A})$ represents the relevance scores for any feature $\textit{A}$, $\textit{diff(A, X, H)}$ is the difference between the values of feature $\textit{A}$ for the two instances $\textit{X}$ and $\textit{H}$, and $\textit{m}$ is the number of instances sampled.

\subsection{Sequential Forward/Backward Selection}
The sequential forward selection (SFS) and sequential backward selection (SBS) are two examples of the wrapper class. SFS \cite{Whitney1971} starts with an empty set and each time it sequentially adds new feature to the set as this feature minimizes the classifier prediction error. Otherwise, it removes this feature and tries another one. It remains in this greedy behavior until the prediction error becomes fixed in certain range. Conversely, SBS \cite{Aha1995} starts with the full set of features and each time it sequentially removes a feature from it as this removal minimizes the classifier prediction error and so on as SFS.

\section{Experimental Results and Analysis}
Three measures are calculated for evaluating the classification performance of selected features subset with ACS. The first measure is the \textit{sensitivity (SN)}, which is the ratio of well-classified vessel pixels. The second measure is the \textit{specificity (SP)}, which is the ratio of well-classified non-vessel pixels. The third measure is the \textit{accuracy (ACC)}, which is the ratio of well-classified vessel and non-vessel pixels. The features subsets resulted via applying the previously stated feature selection heuristics are shown in table 1. 

As shown in table1, there are some features that have been never selected by any heuristic (\textit{$Hu_{6}$},\textit{ $Hu_{7}$} and \textit{$Hu_{8}$}), features that have been selected by minority of heuristics (\textit{Green} gray-level, \textit{$f_{1}$}, \textit{$Hu_{2}$}, \textit{$Hu_{3}$} and \textit{ $Hu_{5}$}), features that have been selected by majority of heuristics (\textit{$f_{3}$}, \textit{$f_{4}$}, \textit{$f_{5}$} and \textit{$Hu_{4}$}), and features that have been selected by all heuristics (\textit{$f_{2}$} and \textit{$Hu_{1}$}). Since \textit{$f_{2}$} and \textit{$Hu_{1}$} features are the most common features between all recommended features subsets, it's strongly indicated that using both of them together can highly discriminate between the vessels and non-vessels pixels. This indication is also emphasized by the last row of table 2, which shows the segmentation performance of each recommended features subset by each heuristic. Fig.1 also depicts the segmentation performance of all recommended features subsets.

\begin{table}
\centering
\caption{Features subset recommended by feature selection heuristics}
{\begin{tabular}{|c|c|c|}
\hline
\textbf{Feature Selection Heuristic}& \textbf{Selected Features Subsets}
\\ \hline
Relief	& $\{f_1, f_2, f_3, f_4, f_5, H_{u1}\}$\\ \hline
CFS	&	$\{f_2, f_3, f_4, f_5, H_{u1}, H_{u4}\}$\\ \hline
Fisher Score & $\{f_2, f_3, H_{u1}, H_{u2}, H_{u3}, H_{u4}\}$\\ \hline
Gini Index  &	$\{f_2, f_4, f_5, H_{u1}, H_{u4}, H_{u5}\}$\\ \hline
SFS  &	 $\{ \textit{Green}gray-level, f_2, f_3, f_5, H_{u1}\}$\\ \hline
SBS  &	 $\{ \textit{Green}gray-level, f_2, f_5, H_{u1}\}$\\ \hline
\end{tabular}
}
\end{table}

\begin{table}
\centering
\caption{Classification performance values of ACS with features subset recommended by feature selection heuristics}
{\begin{tabular}{|c|c|c|c|}
\hline
\textbf{Features Selection Heuristic}& \textbf{SN (\%)}& \textbf{SP (\%)} & \textbf{ACC (\%)}
\\ \hline
Relief &	75.84	&93.88	&91.55\\ \hline
CFS	&75.41	&93.81	&91.43\\ \hline
Fisher Score	&73.88	&93.49&	90.94\\ \hline
Gini Index	&73.50&	93.36	&90.78\\ \hline
SFS	&74.66&	93.42	&91.01\\ \hline
SBS	&74.97&	93.40	&91.04\\ \hline
Common Features {\textit{$f_{5}$} , \textit{$Hu_{1}$}} & 70.78 & 92.65 & 89.86 \\ \hline
\end{tabular}
}
\end{table}

\begin{figure}
\begin{center}
\includegraphics[scale=0.65]{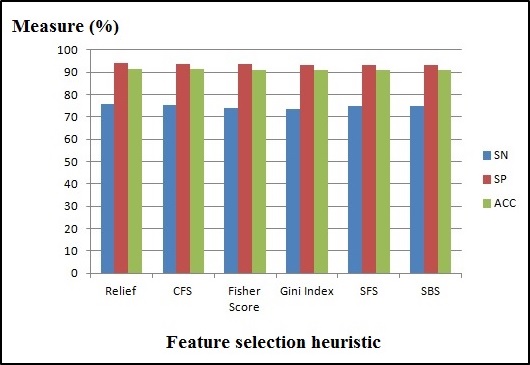}
\\
\caption{Segmentation performance of ACS with the features subset recommended by each heuristic}
\end{center}
\end{figure}

As shown from table 2, there's slight difference between relief and CFS heuristics, which is expected since they intersect in five common features according to table 1. Also, a higher difference between all heuristics is in sensitivity measure than in specificity measure, which means that all recommended features subsets are capable of discriminating vessels pixels more than non-vessels pixels. The current six-feature subset recommended by CFS gives better performance (SN=75.41\%, SP=93.81\% and ACC=91.43\%) than the previous eight-feature subset recommended by CFS, in the previous work \cite{Assad2012} (SN=73.88\%, SP=92.66\% and ACC=90.25\%). That's due to applying different samples selection technique. 

Table 3 shows the computation time of each features subset recommended by each heuristic. As illustrated, there's no significant difference among them and the features computation for each recommended features subset takes about one minute and half on PC with Intel Core-i3 CPU at 2.53 GHz and 3 GB of RAM. 
The features subsets of SFS and SBS takes the lowest computation time, since their features are the fewest and the green gray-level that isn't computed is one of their features. The features subsets of fisher score and gini index takes the longest computation time, because they contain more Hu moment-invariants based features that take more time in computation than gray-level based features and they are computed on two stages. Although this slight difference between these two groups of features in computation, but both take lower computation time when compared against other features used in retinal blood vessels segmentation. That's because both groups are needn't to be computed at multiple scales or orientations.

\begin{table}
\centering
\caption{Computation Time of each features subset recommended by feature selection heuristics}
{\begin{tabular}{|c|c|c|c|c|c|c|}
\hline
\textbf{Features Selection Heuristic}& \textbf{Relief} & \textbf{CFS}& \textbf{Fisher Score} & \textbf{Gini Index} & \textbf{SFS} & \textbf{SBS}
\\ \hline
Computation Time in Seconds	&90	&91&	92&	92 & 89 & 88\\ \hline
\end{tabular}
}
\end{table}

\section{Conclusions and Future Works}

The automatic segmentation of blood vessels is an approach for pixels classification that is based on the computation of many features from the retinal image. The large number of computed features increases the classification complexity and time and reduces its accuracy, so features selection is an essential step for successful classification. This article has evaluated six features selection heuristics to select the best relevant features subset from feature vector consisting of fourteen features. The features are computed for each pixel in the FOV of retinal image in DRIVE database for blood vessels segmentation by ACS. The results indicated that the recommended features subset by the relief heuristic gives the best segmentation performance with sensitivity of 75.84\%, specificity of 93.88\% and accuracy of 91.55\%.

For future work, there are three concurrent directions for improving the accuracy of ACS-based segmentation: the first one is via adding other new features that are as easy to compute as the current features. The second direction is via evaluating other features selection heuristics in order to obtain better features subsets. Finally, the third direction is through applying the segmentation approach, proposed in this article, on other retinal images databases.


\end{document}